%% file: main.tex
\documentclass[runningheads]{llncs}

\usepackage[mobile]{eccv}

\usepackage{eccvabbrv}

\usepackage{graphicx}
\usepackage{booktabs}

\usepackage[accsupp]{axessibility}  

\usepackage[pagebackref,breaklinks,colorlinks,citecolor=eccvblue]{hyperref}

\usepackage{orcidlink}
\usepackage{wrapfig}
\usepackage{subcaption} 

\providecommand{\thetitle}{}
\let\oldtitle\title
\renewcommand{\title}[1]{\oldtitle{#1}\renewcommand{\thetitle}{#1}}
\newcommand{\maketitlesupplementary}{
    \newpage
    \begin{center}
        \Large
        \textbf{\thetitle}\\[0.5em] 
        Supplementary Material\\[1.0em]
    \end{center}
}

\begin{document}

\title{Vista3D: Unravel the 3D Darkside of \\ a Single Image} 

\titlerunning{Vista3D}


\author{Qiuhong Shen\inst{1} \and
Xingyi Yang\inst{1} \and
Michael Bi Mi \inst{2} \and
Xinchao Wang\inst{1}\thanks{Corresponding Author.}\orcidlink{0000-0003-0057-1404}
}

\authorrunning{Shen et al.}

\institute{{\textsuperscript{1} National University of Singapore}
\quad
{\textsuperscript{2} Huawei Technologies Ltd}
\email{\{qiuhong.shen,xyang\}@u.nus.edu} \quad \email{xinchao@nus.edu.sg}
}


\maketitle

\begin{figure}[t]
\centering
\includegraphics[width=1.0\linewidth]{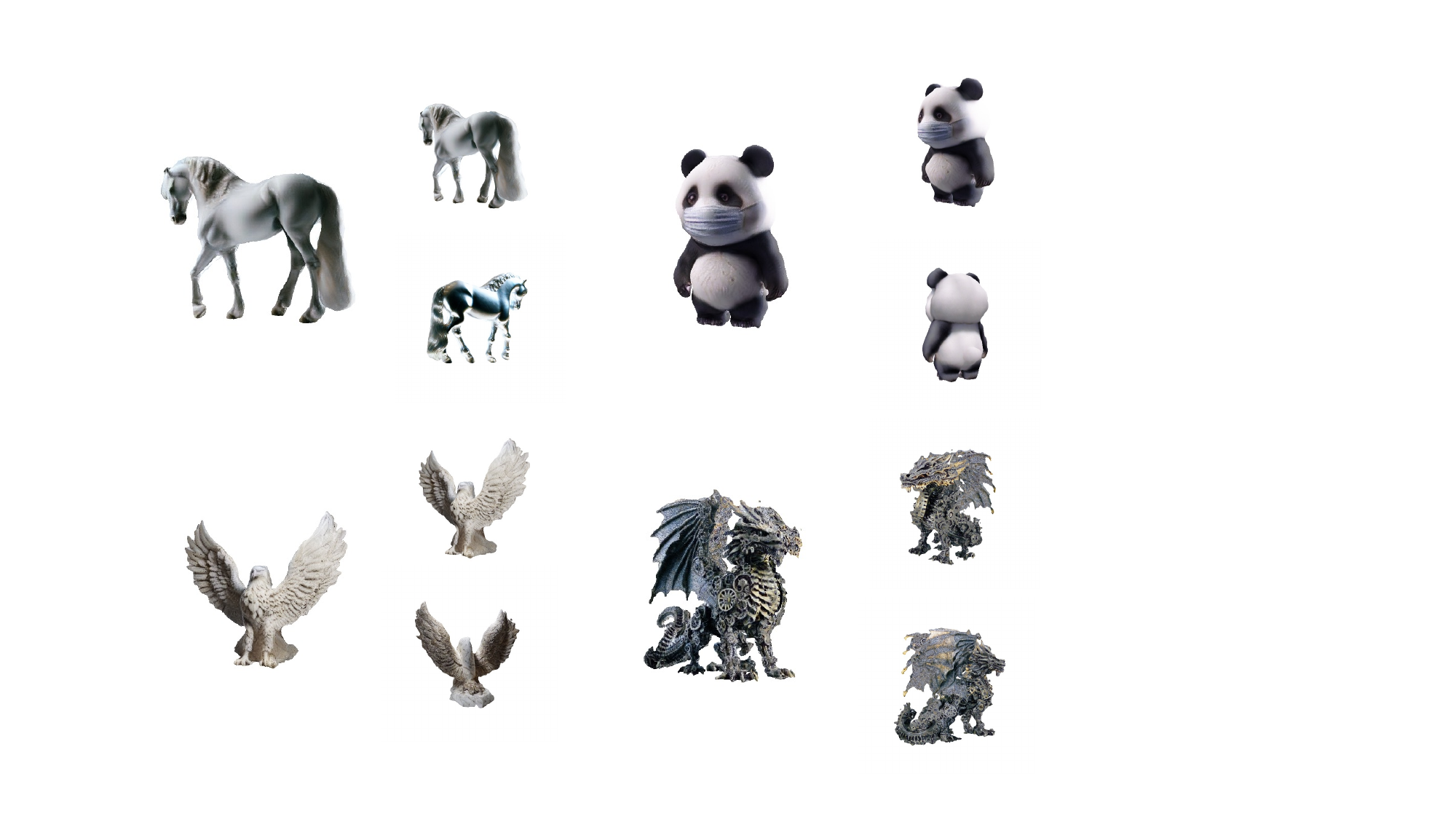}
\caption{\textbf{3D Darkside of Single Image}. By employing various text prompts, Vista3D is capable of unveiling the diversity of unseen views while retaining 3D consistency and detail. Two novel views and the normal map are visualized for each text prompt.}
\label{fig:demo}
\vspace{-6mm}
\end{figure}

\input{sec/0_abstract}    
\input{sec/1_intro}

\input{sec/2_relatedwork}

\input{sec/3_methods}
\input{sec/4_experiments}
\input{sec/5_conclusion}

\input{sec/6_supplementary}


\clearpage  

%
%
\bibliographystyle{splncs04}
\bibliography{main}
\end{document}

%% file: sec/0_abstract.tex
\begin{abstract}
We embark on the age-old quest: unveiling the hidden dimensions of objects from mere glimpses of their visible parts. To address this, we present \textbf{Vista3D}, a framework that realizes swift and consistent 3D generation within a mere 5 minutes. At the heart of Vista3D lies a two-phase approach: the coarse phase and the fine phase. In the coarse phase, we rapidly generate initial geometry with Gaussian Splatting from a single image. In the fine phase, we extract a Signed Distance Function (SDF) directly from learned Gaussian Splatting, optimizing it with a differentiable isosurface representation. Furthermore, it elevates the quality of generation by using a disentangled representation with two independent implicit functions to capture both visible and obscured aspects of objects. Additionally, it harmonizes gradients from 2D diffusion prior with 3D-aware diffusion priors by angular diffusion prior composition. Through extensive evaluation, we demonstrate that Vista3D effectively sustains a balance between the consistency and diversity of the generated 3D objects. Demos and code will be available at \href{https://github.com/florinshen/Vista3D}{https://github.com/florinshen/Vista3D}.

\keywords{3D Generation \and 3D Reconstruction \and Score Distillation}

\vspace{-6mm}
\end{abstract}

%% file: sec/1_intro.tex
\section{Introduction}
\label{sec:intro}
Since the earliest times, our ancestors gazed upon the luminous moon, a symbol of mystery and wonder. Its bright facade, an elegant sphere in the cosmos, has always made us think about what remains hidden: the moon's obscure and elusive dark side. This curiosity, as ancient as human history itself, represents our innate desire to uncover the concealed dimensions that exist beyond the visible. 

This quest, once purely philosophical, has now ventured into the realm of practicality, propelled by the advancements in 3D generative model~\cite{dreamfusion, make3d, anything3d, hash3d, consistent3d}. These technologies enable a broad range of applications, especially in gaming and virtual reality, allowing for the creation of rich, detailed environments and objects without extensive modeling.


Nevertheless, the development of robust large-scale 3D generative models remains a formidable challenge, predominantly due to the limited availability of 3D data. Numerous attempts~\cite{shap-e, point-e, diffTF3D} have been made to train 3D diffusion models on relatively small 3D datasets, condition on textual or visual prompts; Yet, these endeavors often fall short in creating 3D objects with structural integrity and textural consistency.

This challenge is further compounded in the context of reconstructing 3D objects from single images. In this context, two primary approaches emerge.  The first considers the task as a problem of sparse-view reconstruction. However, this often leads to blurred 3D outputs due to the neglect of unseen elements, resulting in excessively blurred 3D objects~\cite{topology3d, pixelnerf} as most views remain unseen. 

On the other hand, the generative approach, which leverages large-scale 2D diffusion models~\cite{dreamfusion, make3d}, introduces its own set of challenges. Efforts to develop 3D-aware 2D diffusion models~\cite{zero123, zero123++, zeronvs, wonder3D, magic123, anything3d, dtc123, dreamcraft3d} involve fine-tuning 2D models with camera transformation modeling on 3D datasets~\cite{objaverse, objaverseXL}. Nevertheless, the prevalence of synthetic objects in these datasets can lead to a compromise in 2D diversity. This often results in the generation of oversimplified geometries and textures.

In this paper, we present Vista3D, a framework designed for reconstructing the unseen view (or "darkside") from a single image. Central to Vista3D is a dual-phase strategy: a coarse phase followed by a fine phase.

In \textbf{the coarse phase}, we leverage 3D Gaussian splatting~\cite{3dgs} to swiftly create basic geometry and textures. To stabilize Gaussian Splatting optimization, we employ a gradient-based Top-K densification strategy, focusing on Gaussian points with the highest gradients.  Additionally, we introduce two novel regularization terms targeting the Gaussian scale and transmittance values, significantly enhancing the convergence speed.

\textbf{The fine phase} then transforms this initial geometry into signed distance fields~(SDF) for further optimization. Here, we employ FlexiCubes~\cite{flexicubes}, an advanced differentiable isosurface technique, to refine the geometry. This refinement aids in learning the signed distance fields (SDFs), deformation, and interpolation weights. The parameters are optimized by ensuring fidelity to the original image and guided by a score function derived from diffusion priors.

Despite these advancements, a unified representation and supervision across all views, both seen and unseen, prove insufficient for capturing the unique characteristics of different viewpoints and generating diverse, consistent 3D objects. To address this, we enhance the representation by implementing \emph{Disentangled Texture Representation}, using two angularly disentangled networks for accurate texture prediction. Furthermore, our \emph{Angular-based Composition} method amalgamates different diffusion priors, adjusting their gradients within specific angular bounds according to their gradient magnitudes. This strategic adjustment assures 3D consistency while promoting diversity in the unseen views. 

Vista3D excels in efficiently generating diverse and consistent 3D objects from a single image within five minutes. Our extensive evaluations demonstrate its ability to maintain a flexible balance between the consistency and diversity of the generated 3D objects.

We summarize our contribution as follows:
\begin{itemize}
\setlength{\itemsep}{0pt}
\setlength{\parsep}{0pt}
\setlength{\parskip}{0pt}

\item We present Vista3D, a framework for revealing the 3D darkside of single images, efficiently generating diverse 3D objects using 2D priors.
\item We develop a transition from Gaussian Splatting to isosurface 3D representations, refining coarse geometry with a differentiable isosurface method and disentangled texture for textured mesh creation.
\item We propose an angular composition approach for diffusion priors, constraining their gradient magnitudes to achieve diversity on the 3D darkside without sacrificing 3D consistency.
\end{itemize}



%% file: sec/2_relatedwork.tex
\section{Related-works}
\label{sec:formatting}


\subsection{3D Generation Conditioned on a Single Image}
The objective of image-to-3D generation is to create 3D objects from a single reference image. Initial methods~\cite{pixelnerf, topology3d} approached this challenge as a variant of sparse view 3D reconstruction. However, these methods often resulted in blurred object outputs due to insufficient priors. Recently, drawing inspiration from text-to-3D initiatives that utilize Score Distillation Sampling (SDS) to elevate 2D diffusion priors into 3D generative models, image-to-3D works~\cite{make3d, anything3d, realfusion, 3dfuse, dreamcraft3d} have adopted a similar approach for 3D object generation based on a single image. However, 2D diffusion priors alone cannot ensure 3D consistency, as they are typically trained solely on image datasets. To address this, several studies~\cite{zero123, zero123++, syncdreamer, wonder3D} have attempted to refine 2D diffusion priors with 3D data~\cite{objaverse, objaverseXL}, enhancing their ability to model 3D consistency. A notable example is Zero-1-to-3, which can generate novel views condition on single image and camera position. Integrating this refined model with SDS~\cite{dreamgaussian, magic123} allows for the reconstruction of coherent 3D objects. 
Moreover, another stream of works~\cite{one2345++, lrm, tgs, gamba, mvgamba, dmv3d, instantmesh} pretrained on large-scale 3D dataset~\cite{objaverseXL} directly predicting the representation of a 3D object from a single image. Diverging from previous works, our work does not solely view this as a 3D reconstruction issue. We redefine it as a 3D generation task aimed at uncovering the unseen 3D aspects behind a single image. Through a meticulously crafted framework, our method efficiently generate diverse and consistent 3D objects.

\subsection{3D Representations for Generation}
Presently, most zero-shot text-to-3D and image-to-3D models utilize an optimization based pipeline, parameterizing the 3D object as a differentiable representation, which varies among different methods. The most prevalent representation in groundbreaking works like dreamfields~\cite{dreamfields}, dreamfusion~\cite{dreamfusion}, and SJC~\cite{sjc} is Neural Radiance Fields (NeRF)~\cite{nerf}. However, training a NeRF is computationally intensive and takes long time to convergence. Magic3D~\cite{magic3d} introduced a two-stage representation, initially learning a coarse NeRF, followed by refining the polygon mesh using a differentiable isosurface method, DMTet~\cite{dmtet}. Fantasia3D~\cite{fantasia3D} suggested directly optimizing DMTet~\cite{dmtet} in separate phases for geometry and texture, but this often leads to mode collapse in the geometry phase and extends training time beyond NeRF. Gaussian Splatting~\cite{3dgs, flashsplat, gs2d, gflow, gof} has gained attention for its efficiency in various 3D tasks, with several 3D generative models~\cite{dreamgaussian, gaussiandreamer, gsgen, luciddreamer} incorporating it for effective generation. However, as a point-based representation, it cannot yield high-fidelity meshes. In our approach, we employ Gaussian Splatting exclusively to create coarse geometry. This coarse geometry is then transformed into SDF, optimized with a hybrid isosurface representation, FlexiCubes~\cite{flexicubes}, to produce high-fidelity meshes. Additionally, we propose an angular disentangled texture representation, tailored to the specifics of this task.

%% file: sec/3_methods.tex
\section{Methodology}
\begin{figure*}[t]
\centering
\includegraphics[width=1.0\linewidth]{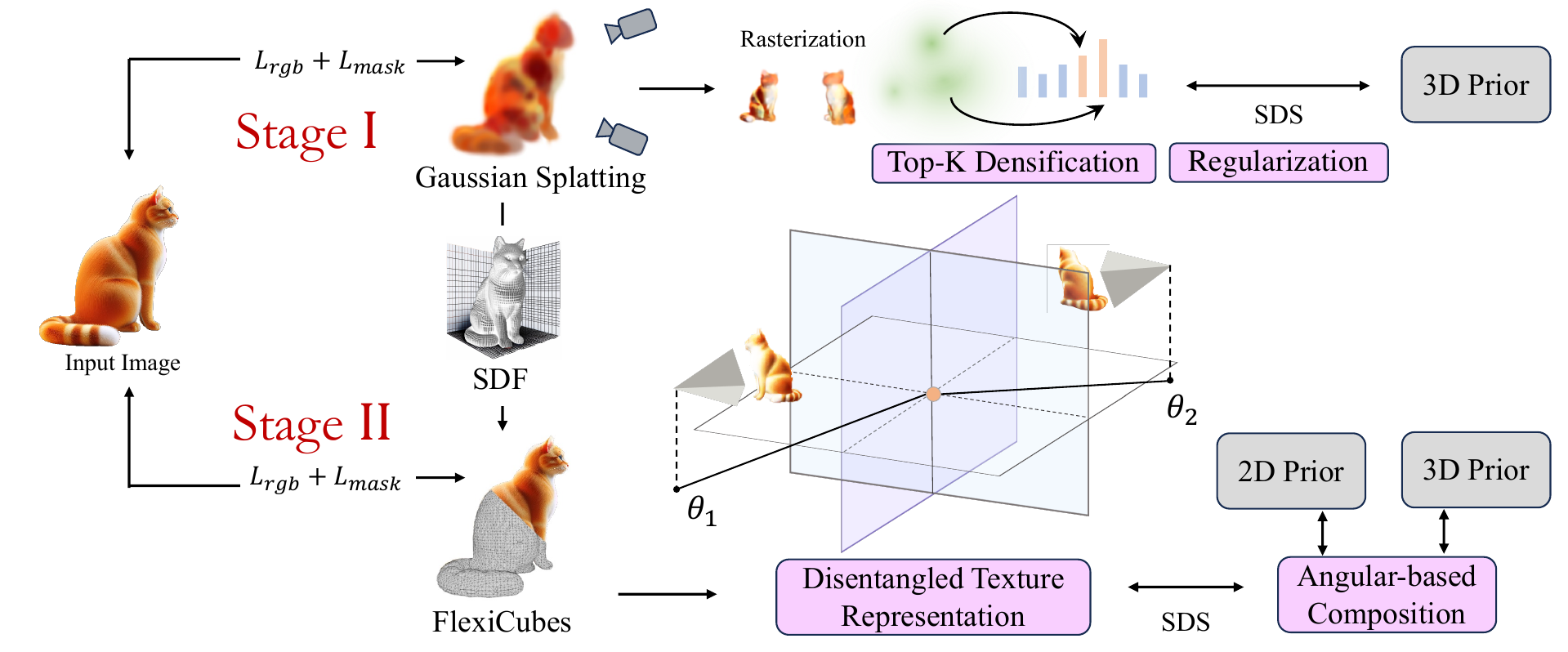}
\caption{\textbf{Overview of Vista3D}. We generate high-fidelity mesh from single image input in a coarse-to-fine manner. In the coarse stage, we utilize Gaussian Splatting to learn a coarse geometry with a 3D-aware 2D diffusion prior. We further extract sign distance fields from Gaussian Splatting for refinement. Another 2D diffusion prior is enabled with an angular-based composition to explore diverse darkside while retain 3D consistency in refinement stage.}
\label{fig:overall}
\vspace{-6mm}
\end{figure*}

\label{sec:method}
In this section, we outline our framework to generate detailed 3D object from single image with 2D diffusion priors. As depicted in Figure~\ref{fig:overall}, our exploration of the 3D darkside of a single image commences with the efficient generation of basic geometry (Section \ref{sec:coarse_stage}), represented through 3D Gaussian Splatting. In refinement stage ({Section \ref{sec:refine_stage}}), we devise a method for transforming the rudimentary 3D Gaussian geometry into signed distance fields, and thereafter, we introduce a differentiable isosurface representation to further enhance the geometry and textures. To enable diverse 3D darkside of given single image, we present a novel approach to constrain two diffusion priors (Section \ref{sec:prior_compose}), enabling the creation of varied yet coherent darkside textures by bounding gradient magnitude. With these approaches, our method can efficiently generate diverse, high-fidelity meshes from a single image.

\subsection{Coarse geometry from Gaussian Splatting}
\label{sec:coarse_stage}

In the coarse stage of our framework, we focus on constructing a basic object geometry using Gaussian Splatting. This technique, as described in~\cite{3dgs}, represents 3D scenes as set of anisotropic 3D Gaussians. 
Compared to other neural inverse rendering methods, such as NeRF~\cite{nerf, instantngp}, Gaussian Splatting demonstrates a notably faster convergence speed in inverse rendering tasks.

Some works~\cite{dreamgaussian, gsgen, gaussiandreamer} has attempted to introduce Gaussian Splatting into 3D generative models. In these methods, we found that directly using Gaussian splatting to generate detailed 3D objects requires optimizing a large number of 3D Gaussians, necessitating significant time for optimization and densification, which is still time-consuming. However, Gaussian Splatting can quickly create a coarse geometry from a single image using a limited number of 3D Gaussians within just one minute. Therefore, in our approach, we utilize Gaussian Splatting solely for the initial coarse geometry generation.

Specifically, each 3D Gaussians is parameterized by its central position $x \in \mathbb{R}^3$, scaling $r \in \mathbb{R}$, rotation quaternion $q \in \mathbb{R}^4$, opacity $\alpha \in \mathbb{R}$, and spherical harmonics $c \in \mathbb{R}^3$ to represent color. To generate a coarse 3D object, we optimize a set of these Gaussian parameters $\Psi = \{ \Phi_{i} \}$, where $\Phi_{i} = \{x_i, r_i, q_i, \alpha_i, c_{i}\}$. To render 3D Gaussians to 2D images, we utilized the highly-optimized tile based rasterization implementation~\cite{3dgs}.

To generate the coarse geometry of given single image $I_{ref}$, we adopt Zero-1-to-3 XL~\cite{zero123, objaverseXL} as 2D diffusion priors $\epsilon_{\phi}$ with pretrained parameters $\phi$. This prior enables denoising of novel views based on the given image $I_{ref}$ and relative camera pose $\Delta \pi$. Accordingly, we optimize the 3D Gaussians $\Psi$ with SDS~\cite{dreamfusion}:
\begin{equation}
\nabla_{\Psi} \mathcal{L}_{SDS} = 
\mathbb{E}_{t,\epsilon} \left[ 
\left( \epsilon_{\phi} \left( I^{\pi}_{R}; t, {I}_{ref}, \Delta \pi \right) 
- \epsilon \right)
\frac{\partial I^{\pi}_{R}}{\partial \Psi} 
\right]
\label{eq:sds}
\end{equation}
where $\pi$ denotes the camera pose sampled around the object with fixed camera radius and $FoV$, $I^{\pi}_{R}$ is the rendered image from 3D Gaussian set $\Psi$ with camera pose $\pi$, timestep $t$ is annealed to weight the gaussian noise $\epsilon$ added to the rendered image. Beyond this basic approach, we introduce a Top-K Gradient-based Densification strategy to accelerate convergence and add two regularization terms to enhance the reconstructed geometry.

\noindent\textbf{Top-K Gradient-based Densification.} In the optimization process, we find the periodical densification~\cite{3dgs} with naive gradient threshold is hard to tune due to the nature randomness of SDS. So we instead use a more robust densification strategy. Only gaussians points with top-k gradients will be densified during each interval, this simple strategy can stablize training cross various given images.

\noindent\textbf{Scale \& Transmittance Regularization.} Additionally, We add two regularization terms to encourage Gaussian Splatting to learn more detailed geometry in this phase. A scale regularization is introduced to avoid too large 3d gaussians, and another transmittance regularization is adopted to encourage the geometry learning from transparent to solid. The overall loss function in this stage can be written as:
\begin{equation}
\begin{aligned}
&\nabla_{\Psi} \mathcal{L}_{\text{coarse}} = \lambda_{SDS} \nabla_{\Psi} \mathcal{L}_{SDS}
+  \lambda_{rgb} \nabla_{\Psi} \mathcal{L}_{rgb}    \\
&\quad + \lambda_{mask} \nabla_{\Psi} \mathcal{L}_{mask} + \underbrace{\lambda_{\text{scale}} \nabla_{\Psi} \sum_i \left\| s_i \right\|}_{\text{Scale Regularization}} \\
& \quad - \underbrace{\lambda_{\text{tr}} \nabla_{\Psi} \text{min}(\tau,  \frac{1}{N_{fg}}\sum_k T_k)}_{\text{Transmittance Regularization}};
\end{aligned}
\label{eq:coarse_loss}
\end{equation}
where $\mathcal{L}_{rgb}$ and $\mathcal{L}_{mask}$ are two MSE loss computed between the rendered reference view and the given image. The term $T_k = \sum_{i} \alpha_i \prod_{j=1}^{i-1} (1 - \alpha_j)$ denotes the transmittance value for the $k$-th pixel in $I_{R}^\pi$, where $N_{fg}$ is the total number of foreground pixels. Additionally, $\tau$ serves as a hyperparameter that is gradually annealed from 0.4 to 0.9, effectively regularizing transmittance over time.

\subsection{Mesh refinement and texture disentanglement}
\label{sec:refine_stage}

In the refinement stage, our focus shifts to transforming the coarse geometry, produced via Gaussian splatting, into signed distance fields (SDF) and refining its parameters using a hybrid representation. 

This stage is crucial for overcoming the challenges presented in the coarse stage, notably the surface artifacts frequently introduced by Gaussian splatting. Due to the inability of Gaussian splatting to provide direct estimates of surface normals, we cannot employ traditional smoothing methods to alleviate these artifacts. To counter this, our method incorporates a hybrid mesh representation, which entails modeling the 3D object's geometry as a differentiable isosurface and learning the texture using two distinct, disentangled networks. This dual approach not only smooths out the surface irregularities but also significantly improves the fidelity and overall quality of the 3D model.

\vspace{2mm}

\noindent{\textbf{Geometry representation.}} We utilize FlexiCubes to represent the geometry in our approach. FlexiCubes is a differentiable isosurface representation which allow local flexible adjustments to the extracted mesh geometry and connectivity~\cite{flexicubes}. The geometry of an object is depicted as a deformable voxel grid with learnable weights. Deformation $\delta_{i} \in \mathbb{R}^3$ and sign distance field (SDF) $s_i \in \mathbb{R}$ is learnt for every vertices $v_i$ in the voxel grid. And interpolation weights $\beta \in \mathbb{R}^{20} $ and splitting weights $\gamma \in \mathbb{R}$ are learnt for each grid cell to position dual vertices and control quadrilaterals splitting. Triangle meshes can be extracted from it differentiablely through Dual Marching Cubes~\cite{dmc}.
\vspace{-1mm}
To bridge the gap between the learned coarse geometry and the isosurface representation, we initially extract a density field from Gaussian splattings using local density queries~\cite{dreamgaussian}, followed by the application of marching cubes~\cite{marchingcubes} to extract a base mesh $M_{coarse}$. Subsequently, we query this base mesh at grid vertices $v_i$ to obtain the initial Signed Distance Field (SDF) $s(v_i)$. For stable optimization, the queried SDF is then scaled as follows:
\begin{equation}
    s(v_i) = \frac{\xi \cdot s(v_i)}{\max\left\{ \left| s_j \right| : s_j \in S, s_j < 0 \right\}}, \,\, \text{where } S = \{s_i\}
\end{equation}
where $s_j < 0$ indicates the field within the object. The scale factor $\xi$ linearly increases from 1 to 3 during the optimization process.

\noindent{\textbf{Disentangled Texture Representation.}} For texture learning, we employ hash encoding followed by a MLP to directly learn albedo. However, distinct from text-to-3D tasks, we recognize two primary supervision sources in this task: the provided reference image and the SDS gradient from 2D Diffusion priors. Typically, a substantial loss weight $\lambda_{rgb}$ is assigned for the reference image. This dominant reference image supervision can decelerate the convergence of textures in unseen views, particularly when unseen views significantly differ from the reference view. 

To address this, we separate the texture into two hash encoding, utilizing a ratio that combines with the relative azimuth angle $\Delta \theta = \theta_{\pi} - \theta_{ref}$, where $\theta_{\pi}$ represents the azimuth of the sampled camera pose $\pi$, and $\theta_{ref}$ is the azimuth of the reference image. The hash encoding for a given query point $\kappa$ in the rasterized triangle mesh is expressed as:
\begin{equation}
E = (1 - \eta)H_{back}(\kappa) + \eta H_{ref}(\kappa)
\label{eq:hashencoding}
\end{equation}
where $H_{ref}$ and $H_{back}$ denote learnable hash encoding facing forward and back, $\eta = (\text{cos}(\Delta \theta) + 1) / 2$ is the balance factor that varies with the sampled azimuth angle. Then the encoded feature $E$ is fed into a MLP predict albedo values. 

\vspace{4mm}
With these geometry and texture representation, we can render the 3D object to images by memory-efficient rasterization coupled with lambertian shading. Above learnable parameters $\Theta$ is refined with $\nabla_{\Theta} \mathcal{L}_{\text{refine}}$:
\begin{equation}
\begin{aligned}
\nabla_{\Theta} \mathcal{L}_{\text{refine}} &= \lambda_{SDS} \nabla_{\Theta} \mathcal{L}_{SDS} \\
& \quad +  \lambda_{SDF} \nabla_{\Theta} \mathcal{L}_{SDF} + \lambda_{\text{consistency}} \nabla_{\Theta} \mathcal{L}_{consistency} \\
& \quad + \lambda_{rgb} \lambda_{SDS} \nabla_{\Theta} \mathcal{L}_{rgb} + \lambda_{mask} \nabla_{\Theta} \mathcal{L}_{mask};
\end{aligned}
\label{eq:refine_loss}
\end{equation}
where the $\mathcal{L}_{SDF}$ is a simple SDF regulariztion term to avoid floaters, $\mathcal{L}_{consistency}$ is a smooth loss applied on surface normals~\cite{realfusion, magic3d}, $\mathcal{L}_{rgb}$ and $\mathcal{L}_{mask}$ are two MSE loss between the rendered reference view and the given image.


\subsection{Darkside Diversity via Prior Composition} 
\label{sec:prior_composition}

In implementing our pipeline, we encountered a key challenge related to the lack of diversity in unseen views. This issue largely stems from the reliance on the Zero-1-to-3 XL prior, a model trained on synthetic 3D objects from Objaverse-XL~\cite{objaverseXL}. While this prior is adept at handling 3D-aware generation based on reference images and relative camera poses, it tends to produce oversimplified or overly smooth results in unseen views. This limitation becomes especially pronounced when dealing with objects captured in the real world.

To address this, we integrate an additional prior from Stable-Diffusion, known for its ability to synthesize diverse images.

\noindent\textbf{Darkside diversification with 2D diffusion.} We introduce a second prior, $\epsilon_{\rho}$ with pretrained parameters $\rho$, leading to two Score Distillation Sampling (SDS) loss terms $\nabla \mathcal{L}_{SDS}^{\phi}$ and $\nabla \mathcal{L}_{SDS}^{\rho}$ (Equation ~\ref{eq:sds}) for optimization. The optimal balance between these two priors remains relatively unexplored. While Magic123\cite{magic123} uses an empirical loss weight of $1/40$ for the latter term, this approach may not fully harness the potential of the 2D prior. The key objective in introducing this 2D prior is to introduce greater diversity in unseen view. A small weight with $\nabla \mathcal{L}_{SDS}^{\rho}$ may largely limit its effect.

To enhance the diversity in the unseen aspects of the given image, we employ a gradient constrain method to merge these two priors. We reformulate the SDS loss as a score function~\cite{dreamfusion}, $\nabla_{\Theta} \mathcal{L}_{SDS}(\phi, \, \mathbf{x}) = - \mathbb{E}_{t, \mathbf{z}_t | \mathbf{x}} \nabla_{\Theta} \text{log} p_{\phi}(\mathbf{z}_t | y)$, where $t$ is the timestep and $z_{t}$ is noise latent.

Here $\nabla \mathcal{L}_{SDS}^{\phi}$ is a 3D-aware term conditioned on $y = \{\Delta \pi, I_{ref}\}$, while $\nabla \mathcal{L}_{SDS}^{\rho}$ is a diverse text-to-image term conditioned on text prompt $y = P_{T}$. With different condition $y$, the score function of these two SDS term varies. To retain 3D consistency of unseen views, the magnitude of $\nabla_{\Theta} \text{log} p_{\rho}(\mathbf{z}_t | y)$ need to be constrained with respect to the 3D-aware term $\nabla_{\Theta} \text{log} p_{\phi}(\mathbf{z}_t | y)$. And to avoid the texture to be over-smoothed by the 3D-aware diffusion model, the magnitude of $\nabla_{\Theta} \text{log} p_{\phi}(\mathbf{z}_t | y)$ is indeed to be constrained with the $\nabla_{\Theta} \text{log} p_{\rho}(\mathbf{z}_t | y)$ term.

\noindent\textbf{Angular-based Score Composition.} Since the noise latents $\mathbf{z}_{t}$ in both priors have different encoding spaces, direct evaluation of their magnitudes using the predicted noise difference $\epsilon_{\rho} - \epsilon$ is not feasible.  Instead, we evaluate the magnitude of these terms by observing their gradient on the rendered image $\mathbf{x}$, specifically $\nabla_{\mathbf{x}} \mathcal{L}_{SDS}$. Consequently, we establish upper and lower bounds for the gradient magnitude ratio of these two SDS terms, allowing for a more accurate and feasible evaluation method:
\begin{equation}
 B_{lower}(\eta, \iota) \leq G = \frac{||\nabla_{\mathbf{x}} \mathcal{L}_{SDS}^{\rho}||_2}{||\nabla_{\mathbf{x}} \mathcal{L}_{SDS}^{\phi}||_2} \leq B_{upper}(\eta, \iota)
 \label{eq:score_composition}
\end{equation}
When this ratio exceeds $B_{upper}$, we adjust the magnitude of $\nabla_{\mathbf{x}} \mathcal{L}_{SDS}^{\rho}$ using the factor $B_{upper} / G$. Conversely, if the ratio falls below $B_{lower}$, we scale the magnitude of $\nabla_{\mathbf{x}} \mathcal{L}_{SDS}^{\phi}$ using $G / B_{lower}$. And this $B_{upper}$ and $B_{lower}$ are regulated by the balance factor $\eta$, influenced by the camera pose, and by iterations $\iota$, facilitating a balance between diversity and 3D consistency.

\label{sec:prior_compose}

%% file: sec/4_experiments.tex
\section{Experiments}

\subsection{Implementation Details}
\noindent{\textbf{Coarse geometry learning}.} 
In this phase, the input image undergoes preprocessing with SAM~\cite{sam, unsegment, anything3d}, where the object is extracted and recentered. We initialize all 3D Gaussians with an opacity of 0.1 and a grey color, confined within a sphere of radius 0.5. The rendering resolution is progressively increased from 64 to 512. This stage involves a total of 500 optimization steps, with the densification and pruning of 3D Gaussians occurring every 100 iterations. The top-K densification starts at a ratio of 0.5 and gradually anneals to 0.1, while the pruning opacity remains constant at 0.1. After the first densification, transmittance regularization is activated and selectively applied to the top-$80\%$ opacity values of 3D Gaussians to avoid affecting transparent Gaussians. Scale regularization is enforced using $L_1$ norm. The weights of $\lambda_{scale}$ and $\lambda_{tr}$ are maintained at 0.01 and 1, respectively, throughout the optimization, whereas $\lambda_{rgb}$ and $\lambda_{mask}$ are gradually increased from 0 to 10000 and 1000, respectively. The timestep for SDS is linearly annealed from 980 to 20. For camera pose sampling, the azimuth is sampled in the range of $[-180, 180]$ and elevation in $[-45, 45]$, with a fixed radius of $r = 2$. This phase of optimizing the coarse geometry takes about 30 s.

\noindent{\textbf{Mesh refinement}.} 
In the refinement phase, we configure the grid size of FlexiCubes to $80^3$ within the space $[-1, 1]^3$. The coarse geometry obtained from the initial stage is recentered and rescaled to initialize the Signed Distance Field (SDF) for the vertices of this grid. Interpolation weights are set to 1, and all deformations start at 0. For texture, we use two hash encodings with a two-layer Multilayer Perceptron (MLP). The batch size is maintained at 4. The learning rate for deformation and interpolation weights is 0.005, while it's 0.001 for SDF, and 0.01 for texture parameters. The rendering resolution is gradually increased from 64 to 512. In Equation~\ref{eq:refine_loss}, the loss weights are set as follows: $\lambda_{rgb}=1500$, $\lambda_{mask}=5000$, $\lambda_{sdf} =1$, and $\lambda_{SDS}=1$. We develop two versions for optimization: \textbf{Vista3D-S} and \textbf{Vista3D-L}. \textbf{Vista3D-S} performs 1000 steps of optimization solely with the 3D-aware prior, aiming to generate 3D mesh within 5 minutes. \textbf{Vista3D-L} undergoes 2000 steps of optimization with two diffusion priors to create more detailed 3D objects. The entire optimization process for Vista3D ranges from 15 to 20 minutes. In this stage, camera poses are sampled using a 3D-aware Gaussian unsampling strategy to expedite convergence (additional details are provided in the supplementary material). All experiments are conducted on an RTX3090 GPU.

\noindent{\textbf{Score distillation sampling}.} 
In SDS optimization, the practice of linearly annealing the timestep $t$ to adjust the noise level has been established as effective for producing higher-quality 3D objects~\cite{dreamtime}. However, in our experiments, we observed that linear annealing may not be the optimal strategy. Consequently, we have implemented an interval annealing approach. In this approach, the timestep $t$ is randomly sampled from an annealing interval rather than adhering to a fixed linear progression. This strategy has been found to effectively mitigate the artifacts commonly observed with linear annealing.

\begin{figure}[h!]
\vspace{-6mm}
\centering
\includegraphics[width=0.94\linewidth]{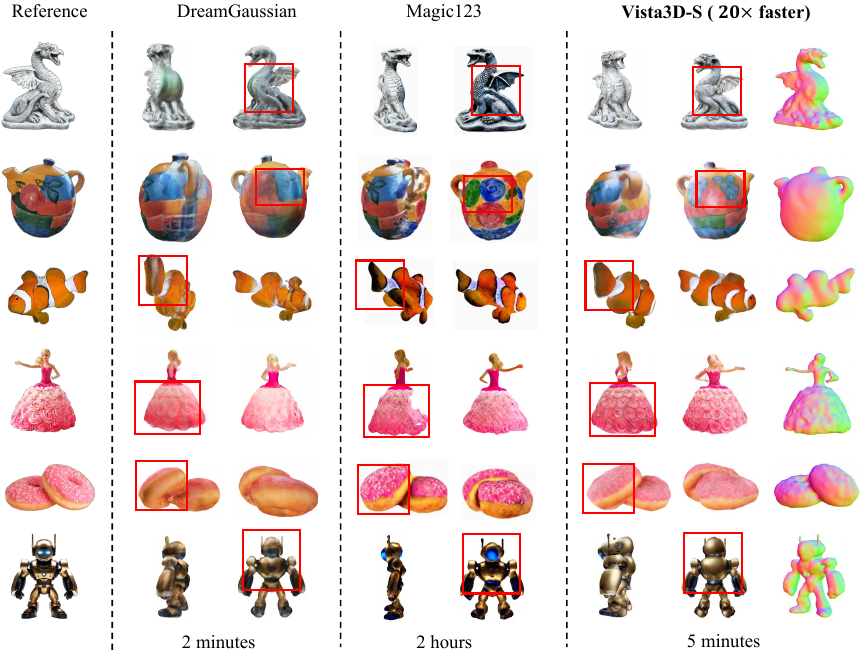}
\caption{\textbf{Qualitative Comparison on image-to-3D generation}. We compare our Vista3D-S with DreamGaussian~\cite{dreamgaussian}, and Magic123~\cite{magic123}. Vista3D-S only takes 5 minutes to reconstruct single 3D object, yielding competitive geometry and more consistent textures compared to Magic123~\cite{magic123} with $20 \times$ speedup.}
\label{fig:sota_comparison}
\vspace{-7mm}
\end{figure}

\noindent{\textbf{Angular diffusion prior composition}.} 
In our model, we utilize two diffusion models: Zero-1-to-3 XL~\cite{zero123, objaverseXL} and the Stable-Diffusion model~\cite{stable-diffusion}. For the Stable-Diffusion model, the timestep $t$ is scaled by the factor $\eta$ to ensure consistency with the reference view. When editing with both diffusion priors, we start with a large initial upper bound $B_{upper} = 100$, which is linearly annealed to 10 across optimization iterations. For front-facing views, where $\eta > 0.75$, we adjust the upper bound using the factor $(1 - \eta)$. The lower bound is specifically implemented for unseen views with $\eta < 0.5$, and its range is gradually reduced from 10 to 1 during the optimization process. For enhancements using the diffusion prior, we apply tighter constraints, with $B_{upper}$ being reduced from 2 to 0.5. The text prompts utilized for the Stable-Diffusion model are derived from the image captions generated by GPT-4.

\begin{figure}[h!]
\vspace{-4mm}
\centering
\includegraphics[width=0.95\linewidth]{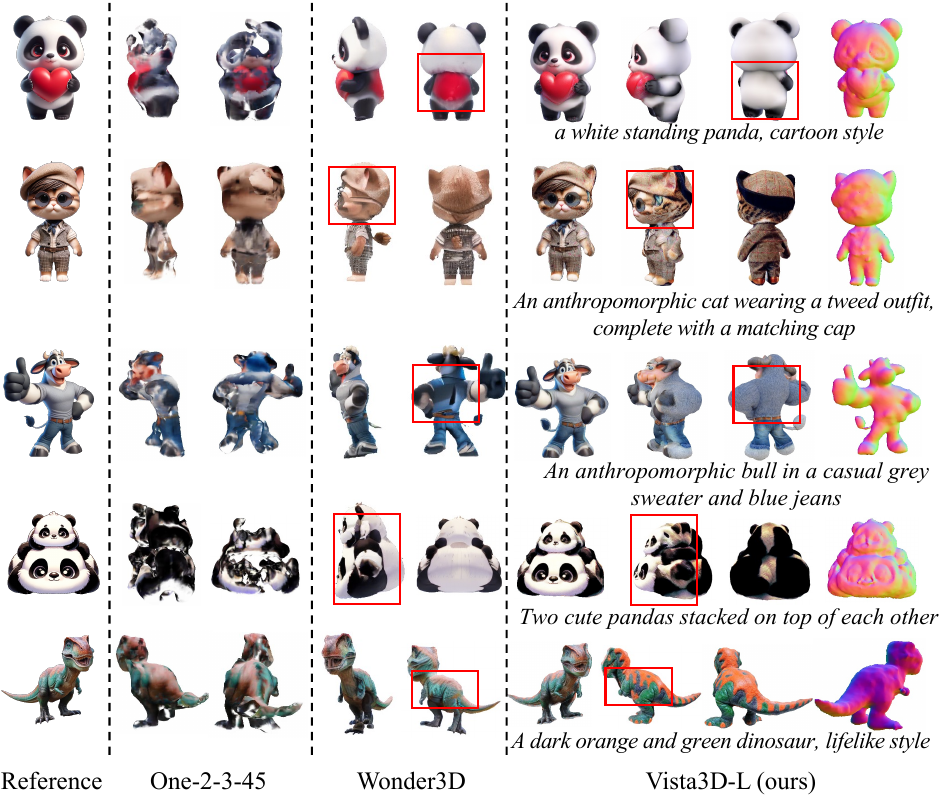}
\caption{\textbf{Qualitative Comparison with One-2-3-45~\cite{one2345} and Wonder3D~\cite{wonder3D}}. In this comparison, we render two views of each 3D object as generated by One-2-3-45 and Wonder3D. For Vista3D-L, we detail the text prompts utilized for the generation of each 3D object, showcasing three rendered views alongside a single normal map for a comprehensive comparison.}
\label{fig:inferonly_compare}
\vspace{-6mm}
\end{figure}

\subsection{Qualitative Comparison}

In Figure~\ref{fig:sota_comparison}, we show our efficient Vista3D-S is capable of generating competitive 3D objects with a $20 \times$ speedup compared to existing coarse-to-fine methods. For Vista3D-L, as depicted in Figure~\ref{fig:demo} and Figure~\ref{fig:inferonly_compare}, we highlight our angular gradient constraint which distinguishes our framework from previous image-to-3D methods, as it can explore the diversity of the backside of single images without sacrificing 3D consistency. In Figure~\ref{fig:sota_comparison}, we primarily compare our Vista3D-S with two baselines, Magic123~\cite{magic123} and DreamGaussian~\cite{dreamgaussian}, for generating 3D objects from a single reference view. Regarding the quality of generated 3D objects, our method outperforms these two methods in terms of both geometry and texture. Regarding Vista3D-L, we compare it with two inference-only single view reconstruction models, specifically One-2-3-45~\cite{one2345} and Wonder3D~\cite{wonder3D}. As shown in Fig.~\ref{fig:inferonly_compare}, One-2-3-45 tends to produce blurred texture and may result in incomplete geometry for more complex objects, while our Vista3D-L achieves more refined textures, particularly on the backside of 3D objects, using user-specified text prompts. And Wonder3D often resorts to simpler textures due to its primary training on synthetic datasets~\cite{objaverseXL}, which occasionally leads to out-of-distribution issues for certain objects. In contrast, Vista3D-L offers zero-shot 3D object reconstruction by controlling two diffusion priors, enabling more detailed and consistent textural. Moreover, given that only a single reference view of the object is provided, we posit that the object should be amenable to editing during optimization with user-specified prompts. To illustrate this, we display several results in Figure~\ref{fig:demo} that emphasize the potential for editing.

\begin{table}[h!]
\vspace{-2mm}
\centering
\resizebox{0.8\linewidth}{!}{\begin{tabular}{c|c|c|c}
\hline
                           & Type    & CLIP-Similarity $\uparrow$ & Time Cost $\downarrow$\\ \hline
One-2-3-45~\cite{one2345}  & Inference     & 0.594           & 45 s      \\
Point-E~\cite{point-e}     & Inference     & 0.587           & 78 s      \\
Shape-E~\cite{shap-e}     & Inference     & 0.591           & 27 s      \\
Zero-1-to-3~\cite{zero123} & Optimization & 0.778           & 30 min    \\
DreamGaussian~\cite{dreamgaussian}  & Optimization & 0.738           & 2 min     \\

Magic123~\cite{magic123}   & Optimization & 0.802           & 2 h   \\ 
DreamCraft3D~\cite{dreamcraft3d} & Optimization & {\color[HTML]{3531FF} 0.842}   & 3.5 h \\ \hline
Vista3D-S                  & Optimization &  0.831           & 5 min     \\
Vista3D-L                  & Optimization & {\color[HTML]{FE0000} 0.868}                & 15 min    \\ \hline
\end{tabular}}
\caption{Quantitative Comparisons on generation quality in terms of CLIP-Similarity for image-to-3D task. Average generation time is reported.}
\label{tab:clip_similarity}
\vspace{-4mm}
\end{table}

\subsection{Quantitative Comparison}
\vspace{-2mm}
In our evaluation, we employ the CLIP-similarity metric~\cite{realfusion, magic123, zero123} to assess the performance of our method in 3D reconstruction using the RealFusion~\cite{realfusion} dataset, which comprises 15 diverse images. Consistent with the settings used in previous studies, we sample 8 views evenly across an azimuth range of $[-180, 180]$ degrees at zero elevation for each object. The cosine similarity is then calculated using the CLIP features of these rendered views and the reference view.
Table~\ref{tab:clip_similarity} highlights that Vista3D-S attains a CLIP-similarity score of 0.831, with an average generation time of just 5 minutes, thereby surpassing the performance of the Magic123~\cite{magic123}. Furthermore, when compared to another optimization-based method, DreamGaussian~\cite{dreamgaussian}, Vista3D-S may take longer at 5 minutes, but it significantly improves consistency, as evidenced by the higher CLIP-Similarity score. For Vista3D-L, we apply an enhancement-only setting. By employing angular diffusion prior composition, our method achieves a higher CLIP-Similarity of 0.868. The capabilities of Vista3D-L, especially in generating objects with more detailed and realistic textures through prior composition, are demonstrated in Figure~\ref{fig:inferonly_compare}.  Additionally, we conduct quantitative experiments on the Google Scanned Object (GSO)~\cite{gso} Dataset, following the setting in SyncDreamer~\cite{syncdreamer}. We evaluate each method using 30 objects and computed PSNR, SSIM, and LPIPS~\cite{lpips} between the rendered views of the 3D object and 16 ground-truth anchor views.
The results, as shown in Tab.~\ref{tab:gso_eval}, reveal that our Vista3D-L achieves SOTA performance among these methods with a large margin. Vista3D-S also demonstrates competitive performance, albeit with a single diffusion prior. 

\begin{table}[h!]
\vspace{-3mm}
\centering
\resizebox{0.6\linewidth}{!}{\begin{tabular}{c|ccc}
\hline
\multicolumn{1}{l|}{} & \multicolumn{1}{l}{PSNR $\uparrow$} & \multicolumn{1}{l}{SSIM $\uparrow$} & \multicolumn{1}{l}{LPIPS $\downarrow$} \\ \hline
RealFusion~\cite{realfusion}     & 15.26                               & 0.722                               & 0.283                                  \\
Make-it-3D~\cite{make3d}     & 15.79                               & 0.741                               & 0.245                                  \\
Zero-1-to-3~\cite{zero123}       & 18.93                               & 0.779                               & 0.166                                  \\
One-2-3-45~\cite{one2345}     & 17.47                               & 0.768                               & 0.184                                  \\
SyncDreamer~\cite{syncdreamer}    & 20.05                               & 0.798                               & 0.146                                  \\
DreamGaussian~\cite{dreamgaussian}  & 23.43                               & 0.832                               & 0.092                                  \\
Magic123~\cite{magic123}      & 24.89                               & 0.875                               & 0.084                                  \\ \hline
Vista3D-S               & 25.42                               & 0.912                               & 0.073                                  \\
Vista3D-L               & {\color[HTML]{FE0000} 26.31}        & {\color[HTML]{FE0000} 0.929}        & {\color[HTML]{FE0000} 0.062}           \\ \hline
\end{tabular}}
\caption{Quantitative Comparison on the GSO~\cite{gso} dataset}
\label{tab:gso_eval}
\end{table}

\subsection{User study}
In our user study, we evaluate reference view consistency and overall 3D model quality~\cite{dreamgaussian}. The evaluation encompasses four methods: DreamGaussian~\cite{dreamgaussian}, Magic123~\cite{magic123}, and our own Vista3D-S and Vista3D-L. We recruited 10 participants for this user study. Each was asked to sort generated 3D object from different methods in terms of view consistency and overall quality respectively. Thus, the scores presented for each metric range from 1 to 4. The results, presented in Table~\ref{tab:user_study}, reveal that our Vista3D-S outperforms the previous methods in both view consistency and overall quality. Furthermore, the adoption of the angular prior composition in Vista3D-L leads to additional improvements in both the consistency and quality of the generated 3D objects.

\begin{table}[h!]
\centering
\setlength{\tabcolsep}{10pt} 
\resizebox{0.9\linewidth}{!}{\begin{tabular}{c|cccc}
\hline
                 & DreamGaussian~\cite{dreamgaussian} & Magic123~\cite{magic123} & Vista3D-S                   & Vista3D-L                   \\ \hline
View Consistency $\uparrow$ & 1.78          & 2.11     & {\color[HTML]{3531FF} 2.87} & {\color[HTML]{FE0000} 3.24} \\
Overall Quality $\uparrow$ & 2.02          & 1.83     & {\color[HTML]{3531FF} 2.81} & {\color[HTML]{FE0000} 3.33} \\ \hline
\end{tabular}}
\caption{\textbf{User study of Vista3D.} We conduct user study in terms of view consistency and overall quality, the score ranges from 1 to 4, the higher the better.}
\label{tab:user_study}
\vspace{-10mm}
\end{table}

\subsection{Ablation Study}

\begin{figure}
\centering
\includegraphics[width=0.94\linewidth]{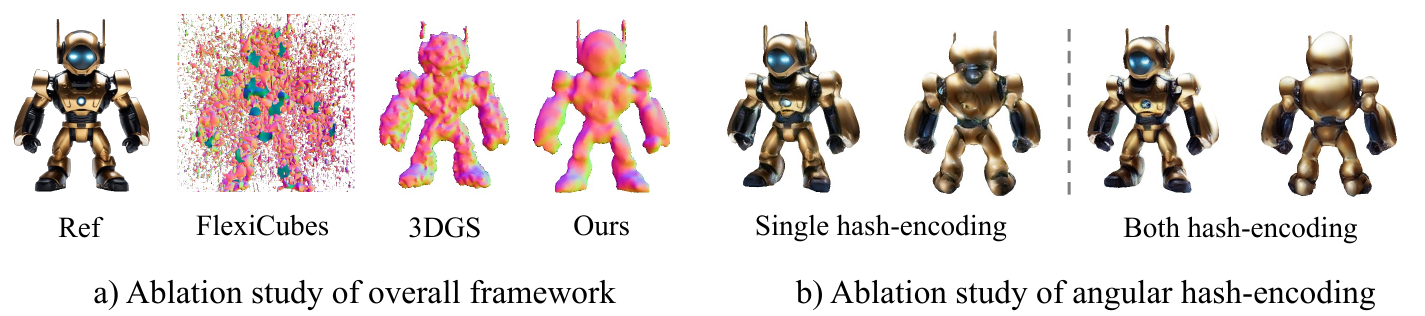}
\caption{\textbf{Ablation study of overall framework and disentangled texture}.}
\label{fig:pipeline_ablation}
\vspace{-4mm}
\end{figure}

\noindent{\textbf{Coarse-to-fine framework.}} Our framework integrates a coarse stage to learn initial geometry then a fine stage to refine geometry and shade textures. We validate the necessity of such a coarse-to-fine pipeline in  Figure~\ref{fig:pipeline_ablation} (a). We first commence with isosurface representation to learn geometry directly, finding the geometry optimization is prone to collapse without preliminary geometry initialization. Thus, a coarse initialization becomes imperative. Beside, we present the normal map of a rough mesh extracted from 3DGS from the coarse stage. It is observed that the coarse stage tends to generate rough even non-watertight geometry, both difficult to mitigate. These findings demonstrate that combining both stages is crucial for the optimal performance of Vista3D.

\noindent{\textbf{Disentangled Texture.}} For validating the effectiveness of the disentangled texture, we compare adopting both hash encodings with single hash encoding in Figure~\ref{fig:pipeline_ablation} (b). With both hash-encodings, the artifacts on the reconstructed robot are notably reduced, especially at the backside. Further, 
we visualize the disentangled texture in supplementary Figure 6(b). Specifically, when visualizing $H_{ref}$, $H_{back}$ is set as $0$ in Equation~\ref{eq:hashencoding}, and vice versa. From the shown visualization, we can clearly find that the facing-forward hash encoding $H_{ref}$ mainly encodes the detail features consistent with the given reference view. While the back hash encoding $H_{back}$ mainly encodes the features in the unseen views. The textures of the facing-forward view and back views are disentangled and learned in two separate hash encodings, which can facilitate learning better textures near the reference view and in unseen views. 


%% file: sec/5_conclusion.tex
\section{Conclusion}
In this paper, we present a coarse-to-fine framework Vista3D to delve into the 3D darkside of a single input image. This framework facilitates user-driven editing through text prompts or enhances generation quality using image captions. The generation process begins with a coarse geometry obtained through Gaussian Splatting, which is subsequently refined using an isosurface representation complemented by disentangled textures. The design of these 3D representations enables the generation of textured meshes within a mere 5 minutes. Additionally, the angular composition of diffusion priors empowers our framework to reveal the diversity of unseen views while maintaining 3D consistency. Our approach surpasses previous methods in terms of realism and detail, striking an optimal balance between generation time and the quality of the textured mesh. We hope our contributions will inspire future advancements and foster future exploration into the 3D darkside of single images.

\section*{Acknowledgement}
This project is supported 
by the Ministry of Education, Singapore, under its Academic Research Fund Tier 2 (Award Number: MOE-T2EP20122-0006),
and the National Research Foundation, Singapore, under its Medium Sized Center for Advanced Robotics Technology Innovation.


%% file: sec/6_supplementary.tex
\clearpage
\setcounter{page}{1}
\setcounter{section}{0}
\maketitlesupplementary

\section{More experimental results}

\subsection{More ablation studies}

\begin{figure}[h!]
    \vspace{-6mm}
    \centering
    \begin{subfigure}[b]{0.49\textwidth}
    \centering
    \includegraphics[width=\textwidth]{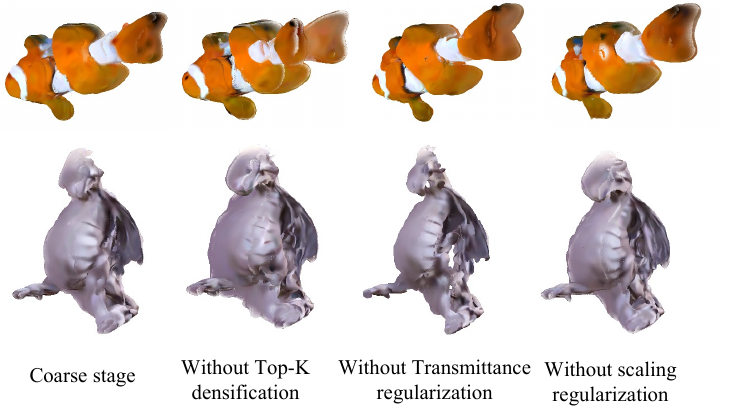}
    \caption{\textbf{Ablation study of the coarse stage}. Here we conduct four settings on the coarse stage, including w/o Top-K densification, w/o transmittance and scaling regularization for comparison.}
    \label{fig:gaussian_ablation}
    \end{subfigure}
    \hfill 
    \centering
    \begin{subfigure}[b]{0.49\textwidth}
    \centering
    \includegraphics[width=\textwidth]{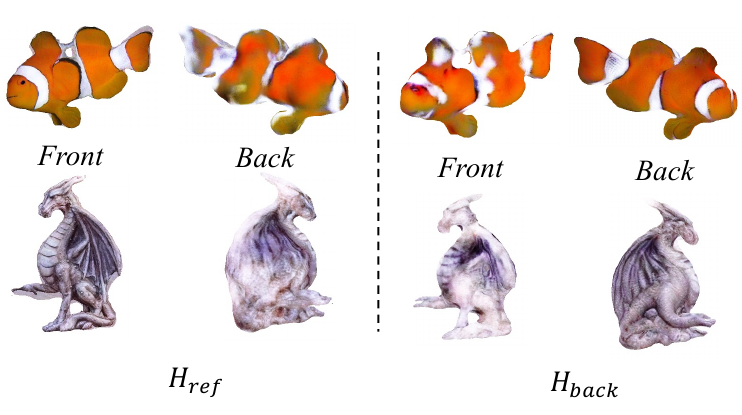}
    \caption{\textbf{Visualization of the disentangled texture}. Here we showcase a generated 3D object. The left side is visualized from the facing-forward hash encoding $H_{ref}$, while the right side is visualized from the back hash encoding $H_{back}$.}
    \label{fig:texture_disen}
    \end{subfigure}
    \hfill
    \caption{Ablation study of the coarse stage and disentangled texture.}
\vspace{-6mm}
\end{figure}

\noindent{\textbf{Top-k densification.}} We compare our densification strategy against a naive gradient threshold approach. This comparison is illustrated in the second column of Figure~\ref{fig:gaussian_ablation}. Using a naive gradient threshold often results in excessive densification of 3D Gaussians, causing geometry to appear swollen. Furthermore, finding an appropriate gradient threshold is challenging, as it varies from case to case. In contrast, our method deterministically controls the densification ratio throughout the optimization process. Consequently, the total number of 3D Gaussians at convergence is solely influenced by the hyperparameter of pruning opacity, effectively maintaining the number of 3D Gaussians within a reasonable range and yielding more accurate geometry.

\noindent{\textbf{Regularization with 3DGS}.} 
In the third and fourth columns of Figure~\ref{fig:gaussian_ablation}, we conduct ablation experiments on the two regularization terms specified in Equation~\ref{eq:coarse_loss}: transmittance regularization and scale regularization. Removing the transmittance regularization tends to produce objects with holes, resulting in coarse meshes from these 3D Gaussians that are often not watertight, complicating refinement stage optimization. On the other hand, excluding only the scale regularization often leads to coarser details in the geometry. This may be caused by Gaussians with larger scales oversmoothing the local geometries.

\begin{figure}[h!]
\setlength{\abovecaptionskip}{0.1cm}
\setlength{\belowcaptionskip}{0cm}
\vspace{-8mm}
\centering
\includegraphics[width=1.0\linewidth]{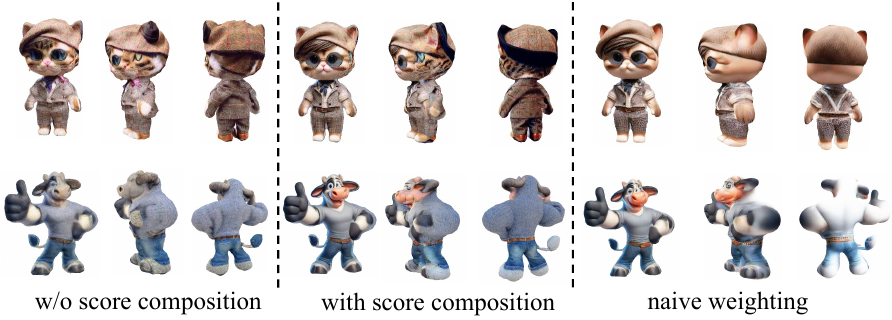}
\caption{\textbf{Ablation Study of Score Composition}. Without score composition, the consistency between the reference view and front view is degraded. Applying naive weighting results in over-smoothed textures on back views.}
\label{fig:prior_composition}
\vspace{-8mm}
\end{figure}

\noindent{\textbf{The effect of prior composition}.} 
To explore the 3D dark side of a single image, we introduce a gradient constraint-based method in Sec.~\ref{sec:prior_composition} to control two diffusion priors in the image-to-3D task. Here we conduct an ablation study to validate the effectiveness of this component. As shown in Fig.~\ref{fig:prior_composition}, without this score composition, though detailed texture on the backside can still be generated, results in degraded consistency between front views and reference images. Another setting involves a naive weighting strategy; we follow Magic123~\cite{magic123} to set a weighting factor of $1/40$ on the SDS term $\mathcal{L}_{SDS}^{\rho}$ with diffusion prior $\epsilon_{\rho}$. With this setting, the backside of the generated 3D objects appears overly smoothed. In contrast, incorporating score composition enables our Vista3D to robustly generate textures that are both detailed and consistent across the front and back views of 3D objects.

\subsection{More qualitative results}

\begin{figure}[h]
\centering
\includegraphics[width=1.0\linewidth]{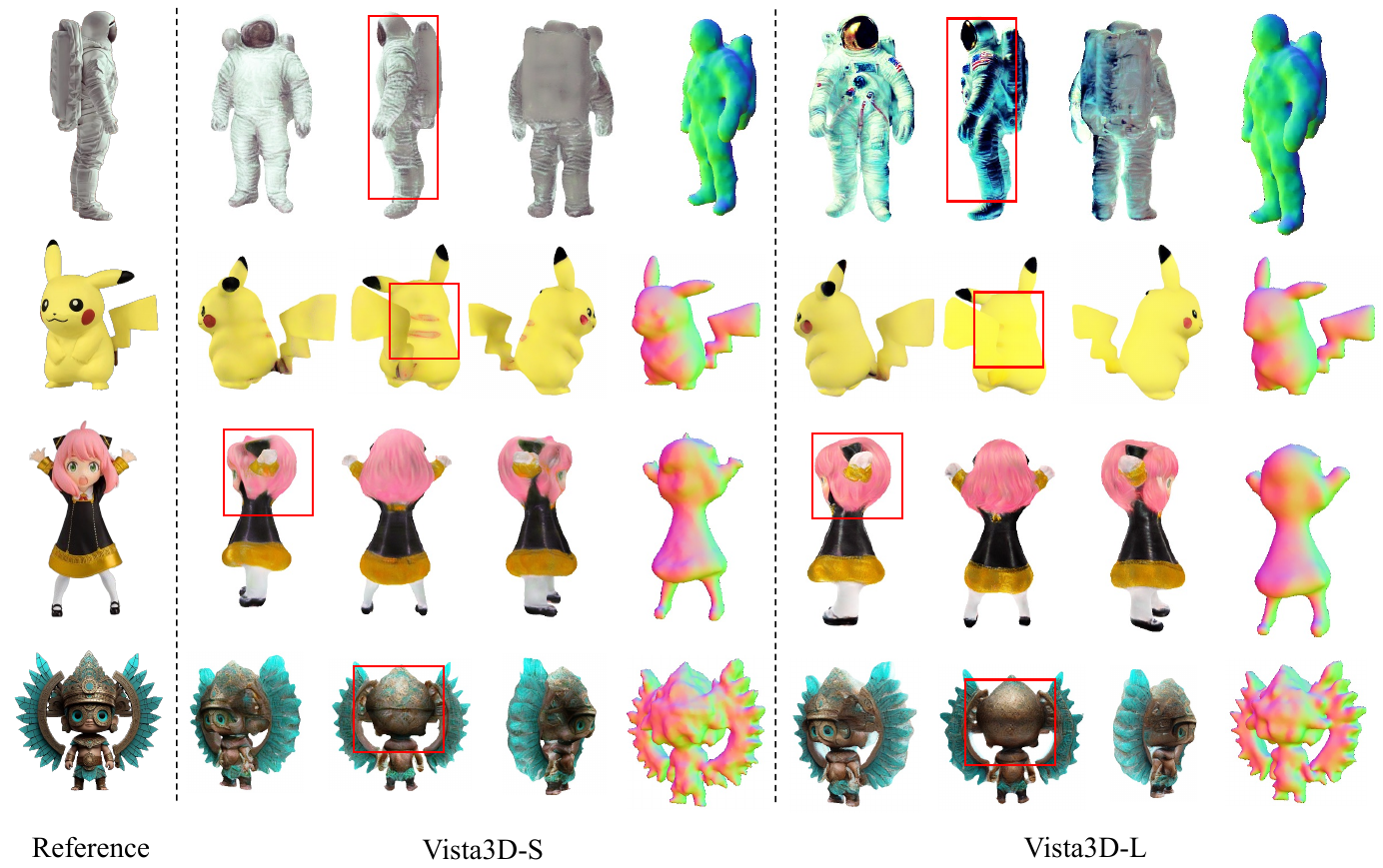}
\caption{Qualitative Comparison between Vista3D-S and Vista3D-L}
\label{fig:only_enhance}
\vspace{-4mm}
\end{figure}

Figure~\ref{fig:only_enhance} showcases the qualitative results of Vista3D-L with diffusion prior composition compared to Vista3D-S with a single diffusion prior. Particularly in scenarios where the provided reference view is less informative, such as when only a side or back view of an object is available, Vista3D-L demonstrates a superior ability to generate more detailed textures compared to Vista3D-S, especially when specific text prompts are used. For example, in the case of the astronaut, Vista3D-S tends to produce oversmoothed textures. In contrast, when using Vista3D-L, the textures generated are notably more vivid and detailed.

\section{Camera Pose Sampling}
\begin{figure}[h!]
\centering
\includegraphics[width=1.0\linewidth]{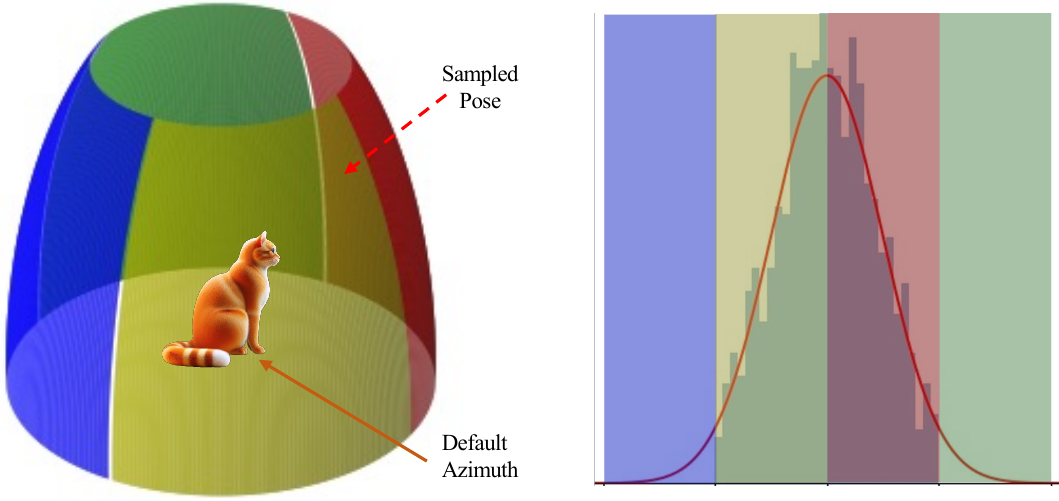}
\caption{\textbf{3D-aware Pose Sampling}, Camera poses are sampled from an empirical PDF with a truncated Gaussian unsampling.}
\label{fig:pose_sample}
\end{figure}

As illustrated in Fig.~\ref{fig:pose_sample}, our approach adopts a 3D-aware camera pose sampling strategy in the refinement stage, diverging from the standard uniform sampling used in previous image-to-3D works~\cite{dreamgaussian, magic123, make3d}. This approach not only speeds up convergence but also enhances visual quality.

Specifically, for a given conditional reference image $I_{ref}$, the pre-trained Zero-1-to-3 model~\cite{zero123} $\epsilon_{\phi}$ is capable of approximating the underlying 3D object distribution $P_{I_{ref}}(x)$. Leveraging this, we employ its estimated empirical error for 3D-aware sampling. 

In this sampling stage, camera poses are sampled from a sphere surface surrounding the central object, divided evenly into $N$ sub-regions ${ R_{i} }$ with azimuth ranging from $[-180, 180]$ degrees, as shown on the left side of Figure~\ref{fig:pose_sample}. Memory queues of fixed length $T$ are established for each sub-region to store empirical errors estimated during the SDS optimization, directly derived from SDS as $(\epsilon_{\phi} - \epsilon)$ in Equation~\ref{eq:sds}. 

When performing pose sampling, an empirical Probability Density Function (PDF) $P_{3d}(R_i)$ is created from these $N$ memory queues. Additionally, given the supplementary supervision from the reference image $I_{ref}$ for forward-facing camera poses, we integrate Gaussian unsampling to reduce sampling frequency on forward-facing poses and increase it for unseen views. This unsampling employs a rejection sampling with a truncated Gaussian distribution, depicted on the right side of Figure~\ref{fig:pose_sample}. Each sub-region is mapped onto this truncated Gaussian PDF, with regions overlapping significantly with the reference view being more likely to be sampled. 

In this process, a camera pose is sampled by initially performing Gaussian unsampling to determine a rejection index $n \in [0, N-1]$. Subsequently, we modify the empirical PDF by setting $P_{3d}(R_{n}) = 0$ and normalizing it. A sub-region index is then sampled from this discrete PDF $\tilde{P}_{3d}(R_i)$, and a camera pose is uniformly sampled from this chosen sub-region.

In our implementation, we configure $N=5$, and initially perform uniform camera pose sampling during the first 100 iterations. For the Gaussian Unsampling, we utilize a truncated Gaussian distribution spanning $[-1, 1]$,  with $\mathcal{N}(0, 0.5)$. This distribution is evenly divided into $N$ intervals to facilitate the sampling process.


\section{Timestep  Sampling in SDS}
Pioneering work DreamFusion~\cite{dreamfusion} randomly sample timestep $t$ from $\mathcal{U} (20, 980)$ in the SDS optimization. However, Dreamtime~\cite{dreamtime} critiques this strategy, suggesting that such random sampling is misaligned with the Denoising Diffusion Probabilistic Models (DDPM) sampling process and leads to inefficient and inaccurate optimization in SDS. Dreamtime suggests a deterministic Time Prioritized (TP) strategy where each iteration step is assigned a unique, decrementally decreasing timestep $t$.

However, we observed that this deterministic approach falls short in SDS optimization. Artifacts generated by large timesteps are not effectively compensated for by smaller timesteps, often exacerbating the problem. To rectify this, we propose an interval-based annealing method for the timestep. Specifically, we define a maximum timestep $t_{max}$ and a minimum timestep $t_{min}$ for each optimization interval, updating them every 50 optimization steps. The timestep is then sampled from the dynamically adjusted interval $\mathcal{U} (t_{min}, t_{max})$. This approach effectively alleviates the artifacts that larger timesteps tend to cause.


\section{Limitations}
Despite Vista3D demonstrating prowess in exploring the 3D dark side of a single image, we acknowledge several limitations for future exploration. Employing a Score Distillation Sampling (SDS) based architecture, Vista3D necessitates optimization for each 3D object it generates, positioning its efficiency a notch below that of purely feed-forward image-to-3D methods. The amount of public 3D data is relatively limited, often resulting in the generation of simplistic 3D objects by feed-forward methodologies. Vista3D leverages diffusion prior composition to facilitate the reconstruction of more diverse 3D objects. This strategy holds promise for the creation of additional 3D data, potentially alleviating the current data scarcity and enabling the development of more sophisticated pretrained image-to-3D models.
\newpage